\documentclass{article}
\usepackage[preprint]{neurips_2021}

\usepackage{natbib} 
\setcitestyle{square,numbers,comma}
\usepackage[utf8]{inputenc} 
\usepackage[T1]{fontenc}    
\usepackage{hyperref}       
\usepackage{url}            
\usepackage{booktabs}       
\usepackage{amsfonts}       
\usepackage{nicefrac}       
\usepackage{microtype}      
\usepackage[pdftex]{graphicx}
\usepackage{xcolor}
\usepackage{subfigure}
\usepackage{wrapfig}
\usepackage{amsmath,bm}
\usepackage{amssymb}
\usepackage{mathtools}
\usepackage{tabularx}
\usepackage{lmodern}
\usepackage{algpseudocode} 
\usepackage[ruled,vlined]{algorithm2e}

\usepackage{amsmath}
\usepackage{cleveref} 
\DeclareMathOperator*{\argmax}{arg\,max}

\DeclarePairedDelimiterX{\infdivx}[2]{\big[}{\big]}{%
  #1\;\delimsize\|\;#2%
}

\newcommand{\E}[1]{\mathop{\mathbb{E}_{#1}}}
\usepackage[colorinlistoftodos]{todonotes}

\usepackage{selectp}

\title{Exploration and preference satisfaction trade-off in reward-free learning}

\author{%
  Noor Sajid\thanks{Corresponding author}\\
  WCHN, University College London, UK \\
  \texttt{noor.sajid.18@ucl.ac.uk} \\
  \And
  Panagiotis Tigas \\
  OATML, Oxford University, UK \\
  \texttt{ptigas@robots.ox.ac.uk} \\
  \And
   Alexey Zakharov \\
   Imperial College London, UK \\
   \texttt{alexey.zakharov19@imperial.ac.uk} \\
  \And
  Zafeirios Fountas \\
  WCHN, University College London, UK \\
  \texttt{zafeirios.fountas@huawei.com} \\
  \And
  Karl Friston \\
  WCHN, University College London, UK \\
  \texttt{k.friston@ucl.ac.uk} \\
}

\begin{document}
\maketitle


\begin{abstract}
Biological agents have meaningful interactions with their environment despite the absence of immediate reward signals. In such instances, the agent can learn preferred modes of behaviour that lead to predictable states -- necessary for survival. In this paper, we pursue the notion that this learnt behaviour can be a consequence of reward-free preference learning that ensures an appropriate trade-off between exploration and preference satisfaction. For this, we introduce a model-based Bayesian agent equipped with a preference learning mechanism (\textbf{pepper}) using conjugate priors. These conjugate priors are used to augment the expected free energy planner for learning preferences over states (or outcomes) across time. Importantly, Pepper enables the agent to learn preferences that encourage adaptive behaviour at test time. We illustrate this in the OpenAI Gym FrozenLake and the 3D mini-world environments -- with and without volatility. Given a constant environment, these agents learn confident (i.e., precise) preferences and act to satisfy them. Conversely, in a volatile setting, perpetual preference uncertainty maintains exploratory behaviour. Our experiments suggest that learnable (reward-free) preferences entail a trade-off between exploration and preference satisfaction. Pepper offers a straightforward framework suitable for designing adaptive agents, when reward functions cannot be predefined as in real  environments. 


\end{abstract}


\section{Introduction}\label{intro}
Extrinsic rewards are not necessary to characterise an agent's interaction with its environment. For example, humans have been shown to rely on intrinsic motivation \cite{ryan2000intrinsic,oudeyer,berlyne1960conflict,wilson2014humans}, that can adequately regulate behaviour\footnote{Explicitly, this prescribes Bayes-optimal behaviour in the sense of Bayesian design and active learning.}. Consequently, in the absence of immediate rewards, there is a preferred exchange with the environment \citep{ashby1961introduction,schrodinger1992life,friston2013life}, that can be updated under changing circumstances. Interestingly, this can result in accruing preferences -- and habits -- that may be at odds with objective goals, e.g., kleptomania. In this paper, we demonstrate that this kind of behaviour can be a consequence of reward-free preference learning that encourages self-evidencing\cite{hohwy2016self} and maintains an appropriate arbitration between exploration and preference satisfaction. In brief, we will see that agents learn to explore or exploit, depending upon the predictability of environmental contingencies.

Preference satisfaction subsumes homeostatic (extrinsic) motivations that encourage individuals to maintain some 'preferred' behaviour \cite{oudeyer} and resist effects of perturbations (external or otherwise). Generally, these refer to base needs that can be satisfied, e.g., going to sleep or eating food. Conversely, exploration involves heterostatic (intrinsic) motivations that distract the agent from its homeostatic imperatives, e.g., novelty-seeking behaviour. Exploratory behaviours would include trying a new hobby or taking a different route to work. This kind of exploration is distinct from random behaviour because it depends upon what the agent does not know. Interestingly, over time, exploration can become the primary mode of behaviour if exploration satisfies the agent’s (learnt) preferences when dealing with an uncertain environment.

Our work is based on the notion that an adaptive agent learns the preferences that best reflect its environment. To this end, we present \textbf{pepper}; a preference learning mechanism that can accumulate preferences over states (or outcomes) using conjugate priors -- given a model-based Bayesian agent. Here, we instantiate Pepper as a deep active inference agent \cite{friston2017active,ueltzhoffer2018deep,catal2020learning,tschantz2020learning,fountas2020deep}, maximising the evidence lower bound (or minimising the free energy) during training, and optimising the expected free energy (or free energy of future trajectories) for planning \cite{parr2019generalised,beren2021}. However, it would be straightforward to add Pepper to other Bayesian reinforcement learning (RL) agents instead e.g., MaxEnt\cite{haarnoja2018learning}, Max\cite{shyam2019model} or Dreamer\cite{hafner2020mastering}, etc. Active inference was chosen deliberately to leverage the expected free energy (EFE) as a planning objective that captures the imperative to maximise: a) intrinsic value -- from interactions with the environment -- about latent states, and b) extrinsic value, namely, realising prior preferences over outcomes. Moreover, active inference's Bayesian formulation provides a natural way to introduce conjugate priors necessary for amortised learning of preferences over states (or outcomes) -- as previously shown in a simplified setting for outcome preference learning\cite{sajid2021active}.

Briefly, Pepper comprises a two-step procedure which occurs after the (generative) model of the agent is optimised for the environment (i.e., training time -- see Fig.~\ref{fig:architecture}). The first step consists of short episodes of direct exchange with the environment, where a history of observations and latent state representations are retained. Once each episode finishes, the second step involves updating prior preferences based on the history using simple update rules (see Section \ref{sec::preflearning}). Importantly, this means that agent can learn (different) preferences that encourage adaptive behaviour at test. 

The key contributions of this work are: %
\vspace{-.7em}
\begin{itemize}
    \itemsep0em 
    \item We present a simple, and flexible, preference learning mechanism (pepper) to augment the planning objective (i.e., EFE) for learning (state or outcome) preferences using deep learning. 
    \item Pepper is reward-free at train and test time. This is achieved by casting rewards as a random variable in our generative model; equivalent to any other observation.
    \item Adaptive behaviour is conceptualised as a trade-off between exploration and preference satisfaction.
\end{itemize}
\vspace{-.7em}
In what follows, we review the related literature. Next, we introduce the problem setting and pepper (the preference learning mechanism). We then evaluate the different types of preferences learnt during test time, and how they engender an appropriate trade-off between exploration and preference satisfaction. Finally, we discuss the potential implications of this work.

\section{Related work}
\vspace{-.5em}
RL is regarded as a suitable framework for building artificial agents. However, by definition, it relies on a reward signal to reinforce agent behaviour \cite{sutton2018reinforcement}. In reality, agents do not operate in a problem-solving setting, where a ``critic'' may not be readily available to provide immediate rewards \cite{barto2004intrinsically,singh2005intrinsically,singh2010intrinsically}. Without task-specific reward signal (also called \textit{extrinsic} reward), the agent is driven by intrinsic motivations that promote exploration, play and curiosity~\cite{ryan2000intrinsic,Singh2004IntrinsicallyMR,singh2005intrinsically}. Over the years, a variety of intrinsic motivation methods have been proposed, largely focusing on exploration, based on information gain \citep{VIME, Still2011AnIA}, prediction error \citep{Achiam2017SurpriseBasedIM, Pathak2017CuriosityDrivenEB, Stadie2015IncentivizingEI}, novelty search  \citep{Lehman2008ExploitingOT, Pranav2019MAX}, curiosity \citep{Jurgen91, Jurgen2007}, entropy \citep{MaxEntropy, SMM}, or empowerment \citep{klyubin2005empowerment,Shakir2015Emp}.

Lately, through the popularisation of self-supervised learning (SSL) methods~\cite{hadsell2006dimensionality,misra2020self}, the deep RL community has turned its attention to self-supervised reinforcement learning. Auxiliary tasks or rewards~\cite{jaderberg2016reinforcement} are used -- in the absence of any extrinsic rewards during train time -- to train intrinsically motivated agents for representation learning~\cite{Oord2018RepresentationLW, Guo2018NeuralPB, Kipf2020ContrastiveLO, Stooke2020DecouplingRL} or generative model learning~\cite{shyam2019model,sekar2020planning,ball2020ready}. Recent work \citep{jin2020reward,wang2020reward} has focused on theoretical properties of reward-free Reinforcement Learning. However, the ultimate goal of such methods is to yield easily transferable representations to be exploited upon introduction of a task.

Our approach differs in several ways. First, we formulate intrinsic motivation using the Expected Free Energy ~\cite{friston2017active,kaplan2018planning} and focus on investigating the behaviour of intrinsically motivated agents. Now exploration is simply an emergent behaviour of the planning objective and not a mechanism for improving future task performance. Second, Pepper can be used to learn preferences over both states and outcomes in a deep learning setting -- extending previous formulations to high-dimensional spaces \cite{sajid2021active}. Conceptually, open-ended learning \cite{schmidhuber2010formal,standish2003open,stanley2017open}, where agents are responsible for never-ending learning opportunities, is closest to our formulation. However, Pepper extends this scenario to show how agents can trade-off between actively looking for opportunities to learn but also enjoy moments of preference satisfaction.

\section{Problem setting}
\begin{figure}[!htp]
  \centering
  \includegraphics[width=\linewidth]{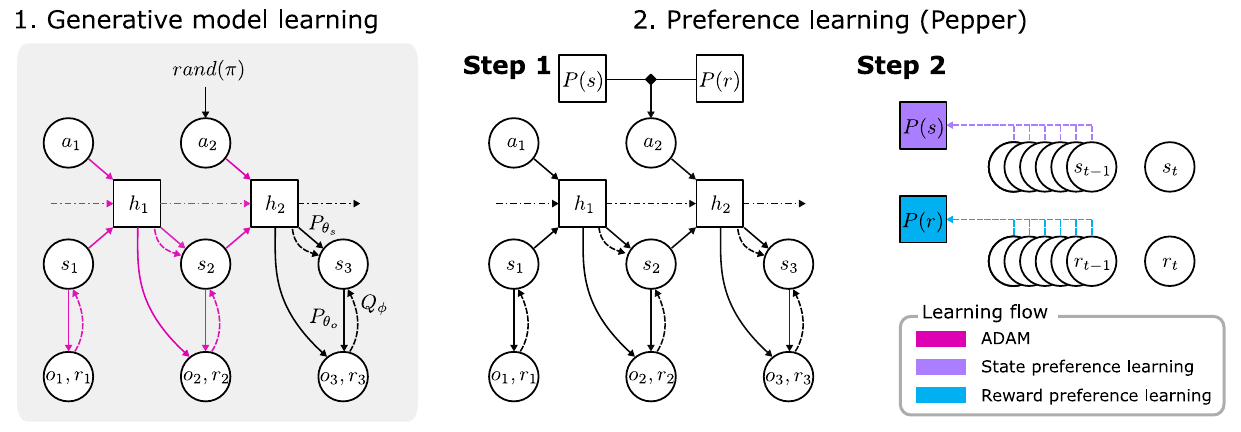}
  \caption{Model architecture and $2$-step training procedure. Circles and squares denote random and deterministic variables respectively. Coloured lines denote connections where learning is employed. Shaded circles represent outcomes that have already been observed by the agent. The first figure shows the generative model used during learning. The second panel is for Pepper (the preference learning phase) -- comprising two steps in each episode: $1$) interaction with the environment, \& $2$) accumulation of preferences once interaction ends. There is a bi-directional flow between the $2$ steps: step $1$ influences preference learning and step $2$ in turn influences environment interaction in the next episode.}
  \label{fig:architecture}
\end{figure}

\vspace{-.5em}
We consider a world that can be represented as a discrete-time Markov decision process (MDP), formally defined as a tuple of finite sets $(\mathcal{S}, \Pi, \Omega, \mathcal{P}, \mathcal{R})$, such that: $s\in \mathcal{S}$ is a particular latent state, $o\in \Omega$ is a particular image observation, $r \in \mathcal{R}$ is a particular reward observation, and $\mathcal{P}$ is a set of transition probabilities. Notice, we cast the reward function as another random variable i.e., no different to an image observation. Further, $\pi \in \Pi$ where $\pi=\{a_1,a_2, ...,a_T\}$ is a policy (i.e., action trajectory) and $\Pi$ a finite set of all possible policies up to a given time horizon $T \in \mathbb{N}^+$ and $\mathbb T= \{0,..,t,..,\tau,T\}$ a finite set which stands for discrete time; $t$ the current time and $\tau$ some future time. In short, we do not assume an optimal state-action policy but consider sequential policy optimisation inherent in active inference. Accordingly, the agent's generative model is defined as a probability density, $P_\theta(o, r, s, \pi )$, parameterised by $\theta$ (Fig.~\ref{fig:architecture}). 

The generative model is instantiated as a Recurrent State-Space Model (RSSM)~\citep{hafner2019learning,hafner2019dream,hafner2020mastering}\footnote{\url{https://github.com/danijar/dreamerv2} (MIT License)} where a history of observations ($o_0, o_1,.., o_t)$ and actions ($a_0, a_1,.., a_t)$ are mapped to a sequence of deterministic states $h_t$. Using these, distributions over the latent states -- both prior and posterior -- can be attained. Formally, this consists of the following:
\vspace{-.7em}
\begin{itemize}
    \item GRU~\cite{cho2014learning} based deterministic recurrent model, $h_t=f_\theta(h_{<t},s_{<t},\pi_{<t})$;
    \item Latent state posterior, $Q_\phi(s_t|h_t,o_t) \sim Cat$, and prior, $P(s) \sim Cat(D)$ 
    \item Transition model, $P_\theta(s_t|h_t) \sim Cat$ ;
    \item Image predictor (or emission model), $Q_\phi(o_t|h_t,s_t) \sim Bernoulli$ ;
    \item Reward model, $Q_\phi(r_t|h_t,s_t) $, and prior, $P(r) \sim Cat(C)$;
\end{itemize}
\vspace{-.7em}
where $Q_\phi(\cdot)$ denotes the approximate distribution, parameterised by $\phi$. 

\subsection{Learning the generative model}
To learn the generative model the evidence lower bound ($ELBO$) of the likelihood $p(o_{1:T}, r_{1:T}\mid \pi)$~\citep{hafner2019learning} or equivalently, the variational free energy \cite{friston2010free,fountas2020deep,sajid2021active} was optimised:
\vspace{-.5em}
\begin{align}\label{e:elbo}
    \mathcal{L}(\theta) = \sum\limits_{t=1}^T \big[& \underbrace{-\mathbb{E}_{Q_\phi}\left[ \ln P_\theta(o_t\mid s_t,\pi) + \ln P_\theta(r_t\mid s_t,\pi) \right]}_{\mbox{reconstruction}}
    + \underbrace{{\mathbb{E}_{Q_\phi}\left[ D_{KL}(Q_\phi \parallel P_\theta(s_t\mid s_{t-1},\pi) \right]}}_{\mbox{dynamics}} \big]
    \vspace{-1em}
\end{align}
where, $D_{KL}$ denotes the Kullback–Leibler divergence. Practically, this entails using trajectories generated under a random policy. See Appendix \ref{appendix::efe} for $ELBO$ implementation.
\vspace{-.5em}
\section{Pepper: preference learning mechanism}\label{sec::preflearning}
After learning the generative model, we substituted the planning objective with the expected free energy. At time-step $t$ and for a time horizon up to time $T$, the expected free energy (EFE) is~\cite{friston2017active,fountas2020deep}:
\vspace{-.8em}
\begin{align}\label{e:efe}
    G(\pi) = \sum_{\tau=t}^T G(\pi, \tau) = \sum_{\tau=t}^T \E{\tilde{Q}} \left[ \log Q_\phi(s_\tau | \pi) - \log \tilde{P}_\theta(o_\tau, s_\tau | \pi) \right] ~ , 
    \vspace{-1em}
\end{align}

where $\tilde{Q} = Q_{\phi}(o_\tau, s_\tau, \theta | \pi) =Q(\theta|\pi)Q(s_{\tau}|,\pi)Q_\phi(o_{\tau}|s_{\tau},\pi)$ and $\tilde{P}_\theta(o_\tau, s_\tau | \pi) = P(o_\tau) Q(s_\tau | o_\tau) P(\theta | s_\tau, o_\tau, \pi)$. This is an appropriate planning objective because it: $1)$ is analogous to the expectation of the $ELBO$ (Eq.\ref{e:elbo}) under the predictive posterior $P_{\theta}(o_t|s_t,\pi)$ and $2)$ can be decomposed into extrinsic and intrinsic value without any additional terms \cite{da2020active,beren2021}. Accordingly, in the absence of learnt preferences -- or whilst learning them -- intrinsic motivation contextualises agent's interactions with the environment environment in a way that depends upon its posterior beliefs about latent environmental states \cite{barto2013intrinsic,ryan2000intrinsic}. 
Actions are selected by sampling from distribution $P(\pi) = \argmax(-G(\pi))$. We Pepper this planning objective with conjugate priors to allow for preference learning over time. 

\subsection{Learning preferences using conjugate priors}
A natural way to learn preferences is to extend the agent's generative model with conjugate priors over prior beliefs (i.e., hyper-priors) for each appropriate probability distributions, that are learnt over time \cite{friston2017active,sajid2021active,da2020active}. Generally, for closed-form updates any exponential family would be appropriate \cite{raiffa1961applied}. Since our distributions of interest, latent state and rewards are Categorical, we used the Dirichlet distribution as the conjugate prior. For the latent state, $P(s)\sim Cat(D)$, this is defined as:
\vspace{-.4em}
\begin{align}
    P(D^{i}|d^{i}) = Dir(d^{i}) \Rightarrow \begin{cases}
      \mathbb{E}_{P(D^{i}|d^{i})} \big[D^i_{ij}\big]=\frac{d^i_{ij}}{\sum_k d^i_{kj}} \\
       \mathbb{E}_{P(D^{i}|d^{i})} \big[log(D^i_{ij})\big]=\digamma(d^i_{ij})-\digamma(\sum_k d^i_{kj})
    \end{cases}  
    \vspace{-.2em}
\end{align}
where, $\digamma$ is the digamma function, $d \sim \mathbb{R}^+$ and same parameterisation holds for $P(r)\sim Cat(C)$. The posteriors for the Dirichlet hyper--parameters are evaluated by updating the prior using the following rule  ${\boldsymbol{d_{i,j}}+\alpha*s_{i,j}}$ where $s_{i,j}$ are the observations for that particular category and $\alpha$ the learning rate. These estimates can be treated as pseudo--counts, and the ensuing learning procedure is reminiscent of Hebbian plasticity -- see \cite{friston2017active,sajid2020degeneracy} for discussion. In effect, this allows the agent to accumulate contingencies and learn about what it prefers -- either via outcomes or states.

\subsubsection{Augmented expected free energy}
To incorporate preference over both states and reward outcomes, we use two distinct EFE decompositions. The first one instantiates preference learning over reward outcomes, as presented below for a single time instance $\tau$ \cite{schwartenbeck2019computational,fountas2020deep}: 
\begin{subequations}\label{eq:G4}
\begin{align}\label{eq:G4a}
G(\pi,\tau) = &-\E{\tilde{Q}}\big[ \log P(r|C) \big] \\\label{eq:G4b}
&+\E{\tilde{Q}}\big[ \log Q(s_{\tau}|\pi) - \log P(s_{\tau}|o_{\tau},\pi) \big]\\\label{eq:G4c}
&+\E{\tilde{Q}}\big[ \log Q(\theta|s_{\tau},\pi) - \log P(\theta|s_{\tau},o_{\tau},\pi) \big] ~ .
\end{align}
\end{subequations}
where, $P(r|C)$ is the probability of a particular reward outcome given (learnt) prior preferences ($C$). The second decomposition incorporates preferences over latent states:
\begin{subequations}\label{eq:G5}
\begin{align}\label{eq:G5a}
G(\pi,\tau) = &-\E{\tilde{Q}}\big[ \log P(o_{\tau}|s_{\tau},\pi) \big] \\\label{eq:G5b}
&+\E{\tilde{Q}}\big[ \log Q(s_{\tau}|\pi) - \log P(s|D) \big]\\\label{eq:G5c}
&+\E{\tilde{Q}}\big[ \log Q(\theta|s_{\tau},\pi) - \log P(\theta|s_{\tau},o_{\tau},\pi) \big] ~ .
\end{align}
\end{subequations}
where, $P(s|D)$ is the probability of a particular state given (learnt) prior preferences ($D$). These prior distributions are read as ‘preferences’ in active inference \cite{sajid2021active} -- in the sense they are the outcomes the agent expects its plans to secure. For both formulations, we drop the conditioning on policy when learning of preferences and the requisite probability is calculated using Thompson sampling. This entails sampling from the prior Dirichlet distribution and estimating the likelihood. Additionally, two of the three terms that constitute the expected free energy cannot be easily computed as written in Eq.\ref{eq:G4} \& Eq.\ref{eq:G5}. To finesse their computation, we re-arrange these expressions and use deep ensembles~\cite{lakshminarayanan2016simple} to render these expressions tractable. See Appendix \ref{appendix::efe} for implementation details. 

\subsubsection{Pepper}
Pepper comprises a double loop during preference learning. The first loop is across time-steps when the agent interacts with the environment. This loop stores information about what happened (including observations, rewards, posterior, prior, etc). The second loop, evolving at a slower timescale, is across episodes and entails preference learning. Specifically, once the interaction with the environment ends, the agent updates the prior Dirichlet distribution (over preferences) using the data gathered during the episode. In the subsequent time-steps the updated preferences are used to select the next action (via their influence on expected free energy). The data gathered during this episode are used to update preferences. In turn, these preferences are used to select actions during the next episode, and so on. This demonstrates a bi-directional flow between the two loops: information from environment interactions influences preference learning and the learnt preferences influence environment interaction in the subsequent episode. The pepper preference learning procedure is summarised in Algorithm.\ref{pseudocode}.

\begin{algorithm}[H]
\label{pseudocode}  
 \caption{Pepper}
\SetEndCharOfAlgoLine{}
 \SetKwComment{Comment}{// }{}
 \SetKwInOut{Input}{Input}
 \Input{\\\hspace{-3.6em}\small
    \begin{tabular}[t]{l @{\hspace{.25em}} l}%
        $h_t := f_\theta(h_{<t},s_{<t},\pi_{<t})$ & Recurrent model\\
        $Q_\phi(s_t|h_t,o_t)$ & Posterior model\\
        $Q_\phi(s_t|h_t)$ & Prior model\\
        $P_\theta(o_t|h_t,s_t)$ & Observation model\\
        $P_\theta(r_t|h_t,s_t)$ & Reward model
    \end{tabular}%
  }
 \textbf{Initialise} \;
 uniform Dirichlet prior over $P(r)$ or $P(s)$ \tcc*[r]{prior preference being learnt}
 learning rate $\alpha$ \;
 \For{\textup{each episode} e}{
     reset environment and collect initial observations ($o_0$ or $r_0$)\;
     \For{ \textup{each time step} t}{
      compute $s_{po} \sim Q_\phi(s_t|h_t,o_t)$ or $s_{pr} \sim Q_\phi(s_t|h_t)$ \;
      compute G (Eq.\ref{eq:G4} or Eq.\ref{eq:G5}) using (learnt) priors, observed and predicted posteriors \;
      $a_t \leftarrow \argmax(-G(\pi)$) \;
      \textbf{execute} ~ $a_t$ and receive $o$ or $r$ \;
      $o_{t+1} \leftarrow o$ and $r_{t+1} \leftarrow r$ \;
     }
    
    \uIf{\textup{reward preference learning}}{
        $c_{i,t} \leftarrow c_{i,t-1} + \alpha*r_{i,t}$  \tcc*[r]{Update rule for $dir(c)$}
    }
    \uElseIf{\textup{state preference learning}}{
        $d_{ij,t} \leftarrow d_{ij,t-1} + \alpha*s_{pr_{ij,t}}$ \tcc*[r]{Update rule for $dir(d)$}
    }
    }
\end{algorithm}
\section{Results}
Here, we present two sets of numerical experiments that underwrite the face validity of pepper in two and three-dimensional environments, respectively.

\begin{figure}[!htbp]
  \centering
  \includegraphics[width=\linewidth]{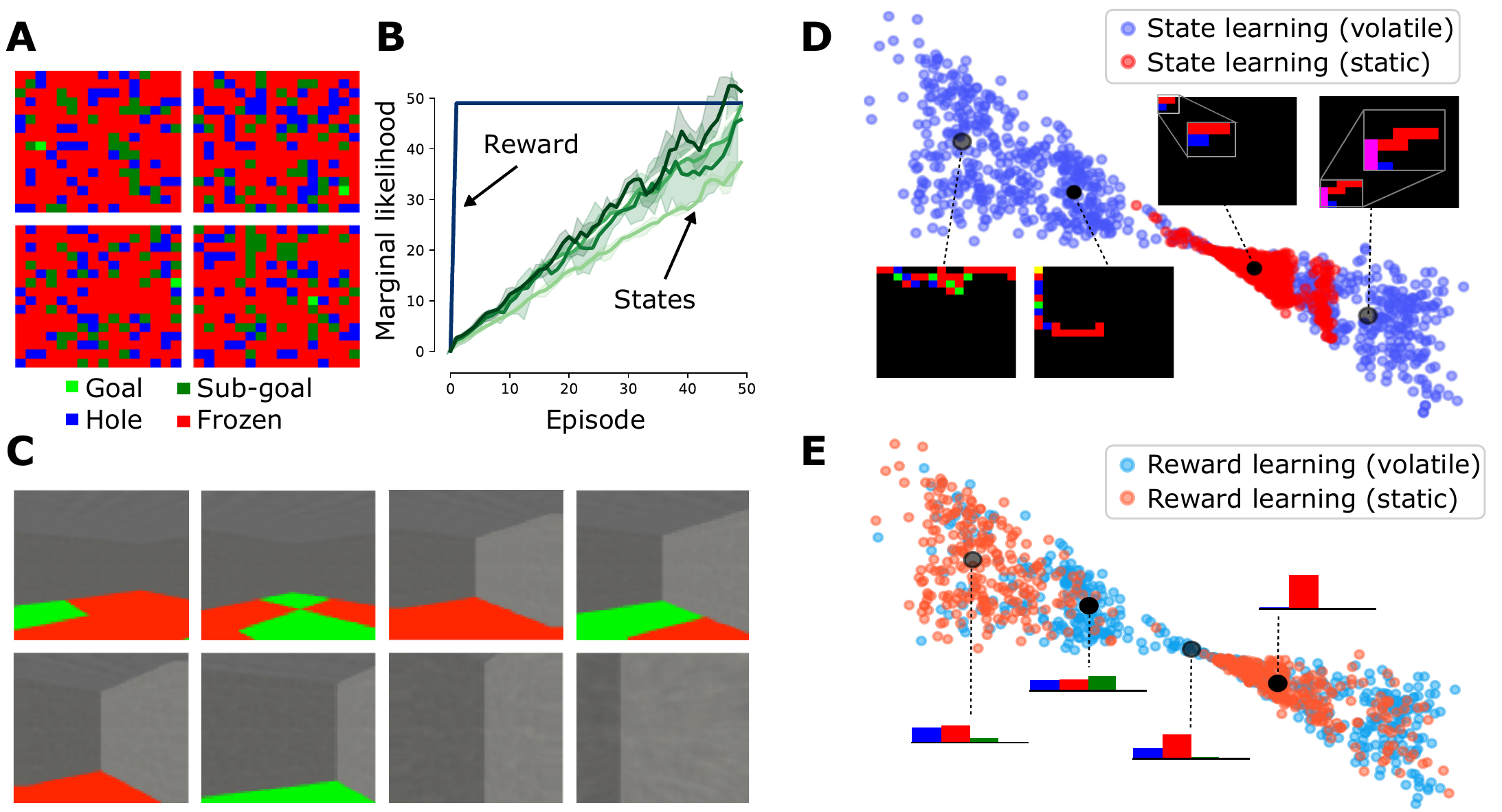}
  \caption{\textbf{A:} The graphics show examples of the different OpenAI Gym FrozenLake 16$x$16 environments used. \textbf{B:} The line plots shows the marginal likelihood (y-axis), across $50$ preference learning episodes (x-axis), for $P(s)$ (i.e., states) and $P(r)$ (i.e., reward) during test time. Here, the dark lines represent the mean (across $10$ seeds), and shaded area the $95\%$ confidence interval. Different shades of green denote levels of environment volatility. \textbf{C:} The graphics show a particular agent trajectory across the $3D-$ TileWorld environment -- with a $10-$step interval between each image \textbf{D-E:} Visualisation of the posterior latent states (estimated using $Q_\phi(s_t|h_t,o_t)$) during preference learning of ($P(s)$ \textbf{D} and $P(r)$ \textbf{E}) episodes. The states have been projected onto the first two principle components, and the black circles represent their k-mean centroid. Here, the accompanying graphics present a representative agent trajectory (\textbf{D} with visited tiles highlighted) and reward profile (\textbf{E}) from that particular cluster. 
  } 
  \label{fig:learnpreferences} 
\end{figure}
\vspace{-.5em}
\subsection{FrozenLake}
We used a variation of the OpenAI Gym~\cite{brockman2016openai}\footnote{\url{https://github.com/openai/gym/} (MIT license)} FrozenLake environment to: $1)$ evaluate different behaviours acquired (at test time) when either $P(s)$ or $P(r)$ was learnt, $2)$ qualify how preferences can evolve as a result of environmental volatility, and $3)$ quantify the trade-off between exploration and preference satisfaction. The agent in the original FrozenLake formulation is tasked with navigating a grid world comprised of frozen, hole and goal tiles, using $4$ actions (left, right, down or up). The agent receives a reward of $10$ upon reaching the goal and a penalty $-0.25$ upon moving to the hole. To test preference learning, we included a sub-goal tile and removed the reward signal (Fig.\ref{fig:learnpreferences}A). In other words, although the preference learning agent can differentiate between tile categories -- given its generative model -- it receives no extrinsic signal from the environment. Here, we simulated a volatile environment by switching the FrozenLake tile configuration every $K$ steps and initialising the agent in a different location at the start of each episode. This furnished an appropriate test-bed to assess how much volatility was necessary to induce exploratory behaviour and shifts in learnt prior preference.

\subsection{TileWorld}
We extended the FrozenLake environment in the miniworld framework~\cite{gym_miniworld}\footnote{\url{https://github.com/maximecb/gym-miniworld/} (Apache 2.0 License)} to three dimensions to test the generalisation and scalability of preference learning, when operating in a 3D visual world  (Fig.\ref{fig:learnpreferences}C). In this task, the agent moves around in a small room with grey walls, frozen and goal tiles on the floor. The agent spawns in a random location and receives pixel observations (32x32 pixels with RGB channels) and a scalar value (reward) containing some information regarding the tile its currently on ($1$ for red tiles, $2$ for green tiles). Additionally, we introduce environmental volatility by changing the floor tiles to a random map and back to the original map every $K$ steps. Alternating between the original and a random map every $K$ step is important to promote exploratory, novelty-seeking behaviour.

In the experiments that follow, we test agent behaviour in the two environments, with and without volatility. The Dirichlet distribution for either prior preference distribution, $P(s)$ or $P(r)$, was initialised as 1 (i.e., uniform preferences). Trained network weights, optimised Eq.\ref{e:elbo} using ADAM \cite{kingma2014adam}, were frozen during these experiments. Therefore, behavioural differences are a direct consequence of pepper that induces differences in estimation of the EFE. See Appendix \ref{appendix::traingenmodel} for architecture and training details for each environment. 

\subsection{Learnt preferences}
\begin{figure}[!tp]
  \centering
  \includegraphics[width=\linewidth]{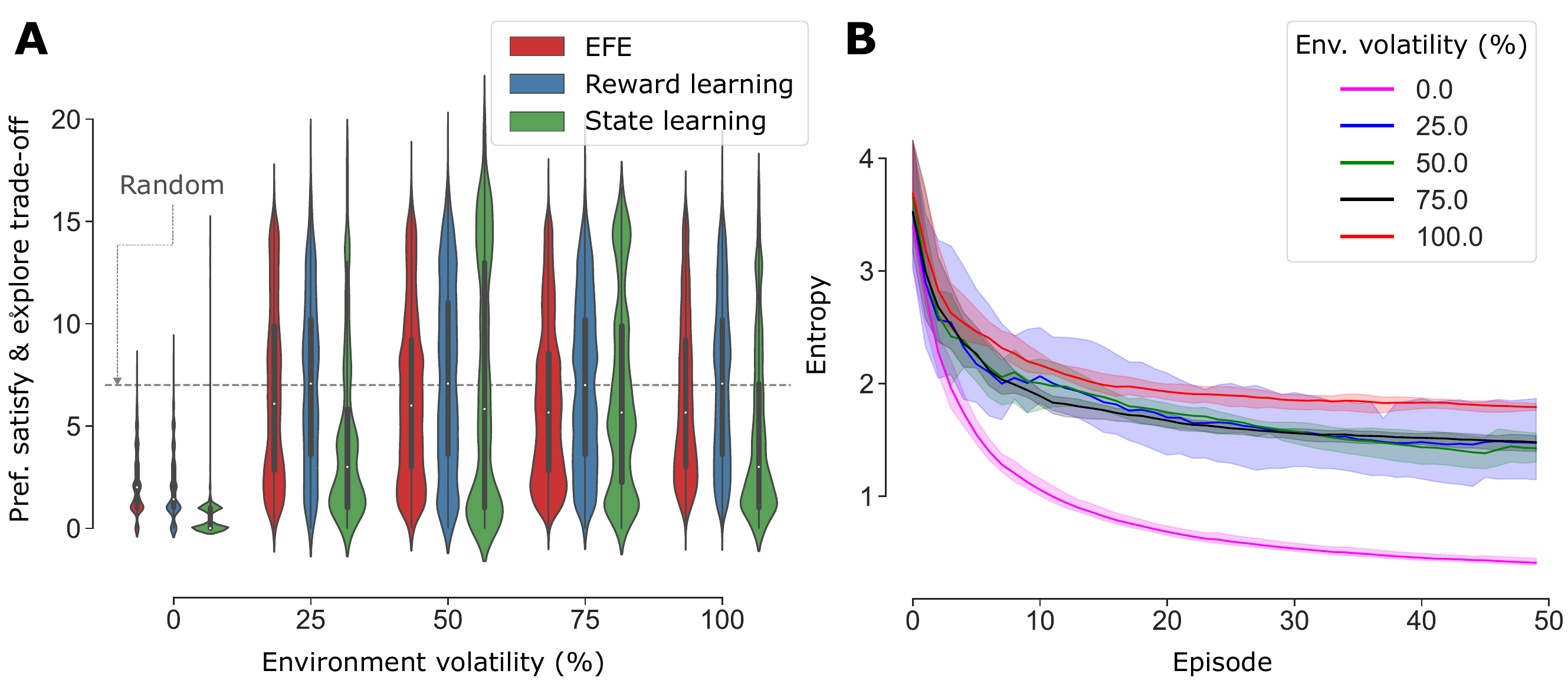}
  \caption{\textbf{A:} The violin plot presents the preference satisfaction and exploration trade-off measured using Hausdorff distance \cite{blumberg1920hausdorff} at different levels of volatility in the environment. The x-axis denotes environment volatility at constant map ($0\%$), change in map every $40$ steps ($25\%$), $20$ steps ($50\%$), $10$ steps ($75\%$) and every step ($100\%$). The y-axis denotes the Hausdorff distance. Here, red is for the agent optimising the standard expected free energy (EFE) Eq.\ref{e:efe}, blue for reward preference learning Eq.\ref{eq:G4} and green for state preference learning Eq.\ref{eq:G5}. \textbf{B:} The line plot depicts the entropy over $P(s)$ across varying levels of volatility in the environment. The x-axis represents the episodes, and the y-axis entropy (in natural units). Here, the dark lines represent the mean (across $10$ seeds), and shaded area the $95\%$ confidence interval. The pink line is for $0\%$, blue for $25\%$, green for $50\%$, black for $75\%$ and red for $100\%$ volatility in the environment.}
  \label{fig:tradeoff} 
\end{figure}

\paragraph{State preferences} Unsurprisingly, preference learning over latent states in the FrozenLake environment revealed two types of behaviours: exploration and preference satisfaction (Fig.\ref{fig:learnpreferences}D $\&$ \ref{fig:tradeoff}A). Here, under a static setting, preference satisfaction entailed restricted movement within a small section of the FrozenLake with gradual accumulation of prior preferences (see example trajectories in Appendix \ref{appendix::preferences}). This speaks to the self-evidencing nature of Pepper. That is as the Pepper agent sees similar observations across episodes it grows increasingly confident (via increased precision over the prior preference) that these are the outcomes it prefers. Conversely, exploratory behaviour is evident in a volatile setting (Fig.\ref{fig:learnpreferences}D $\&$ \ref{fig:tradeoff}A), as gradual preference accumulation entails encoding of previously unseen states (see example trajectories in Appendix \ref{appendix::preferences}). 

\vspace{-1em}
\paragraph{Reward preferences} Reward preference learning revealed subtle differences in preferences (Fig.\ref{fig:learnpreferences}E $\&$ \ref{fig:tradeoff}A), where certain agents preferred sub-goal tiles more than neutral tiles. All Pepper agents were able to immediately maximise their marginal likelihood\footnote{Marginal likelihood is simply the likelihood function of the parameter of interest (here state or reward) where some parameter variables have been marginalised out.} over the reward (Fig.\ref{fig:learnpreferences}B). However we did not observe clear differences in preference learning as the environment context shifted from non-volatile (i.e., map change) to increasingly volatile (map changed every time step). We consider this to be consequence of the sparse categories over the reward distribution, and the large percentage of map being taken up by the neutral tiles. This meant preference accumulation was biased in favour of the neutral tile (see examples in Appendix \ref{appendix::preferences}). 

\subsection{Exploration and preference satisfaction trade-off}
We evaluated the exploration and preference satisfaction trade-off using Hausdorff distance (Fig.\ref{fig:tradeoff}) \cite{blumberg1920hausdorff}. This is an appropriate metric, which calculates the maximum distance of the agents position in a particular trajectory to the nearest position taken in another trajectory. Accordingly, a high Hausdorff distance denotes increased exploration, since trajectories observed across episodes differ from one other. Whereas, a low distance entails prior preference satisfaction as agents repeat trajectories across episodes. Using this metric, an inverted u-shaped association (Fig.\ref{fig:tradeoff}A), between volatility in the environment and preference satisfaction, is observed for preference learning over the states. Here, $50\%$ volatility in the environment, shifts the pepper agents behaviour from satisfying preferences to becoming exploratory, when faced with an uncertain environment (and inability to predict the future). Agents in this setting (with the highest Hausdorff distance) tend to pursue long paths from the initial location. (Fig.\ref{fig:learnpreferences}D).   

Interestingly, at $100\%$ environmental volatility, the Pepper agents behaviour shifts back to satisfying its preferences. These agents learn bi-modal preferences over the latent states. Regardless of how the map changes they move directly to either location given the initial position (see trajectories in Appendix \ref{appendix::exploration}). This ability to disregard random, noisy information with continuous map changes (analogous to the noisy-TV setting introduced in \cite{burda2018exploration}) highlights a motivation beyond random exploration under state preference learning. For reward learning, exploration is also instantiated with increased volatility in the environment. Yet, complete volatility does not trigger a definitive shift back to preference satisfaction. Additionally, the expected free energy agent, without preference learning capacity, also exhibits a shift in behaviour as the environment becomes volatile. The exploratory behaviour here is driven exclusively by an imperative to resolve state uncertainty, (Eq.\ref{eq:G4b}): i.e., the mutual information between the agent’s beliefs about its latent state representation of the world, before and after making a new observation.

\subsection{Preference learning in the volatile TileWorld}
\begin{figure}[!tp]
  \centering
  \includegraphics[width=\linewidth]{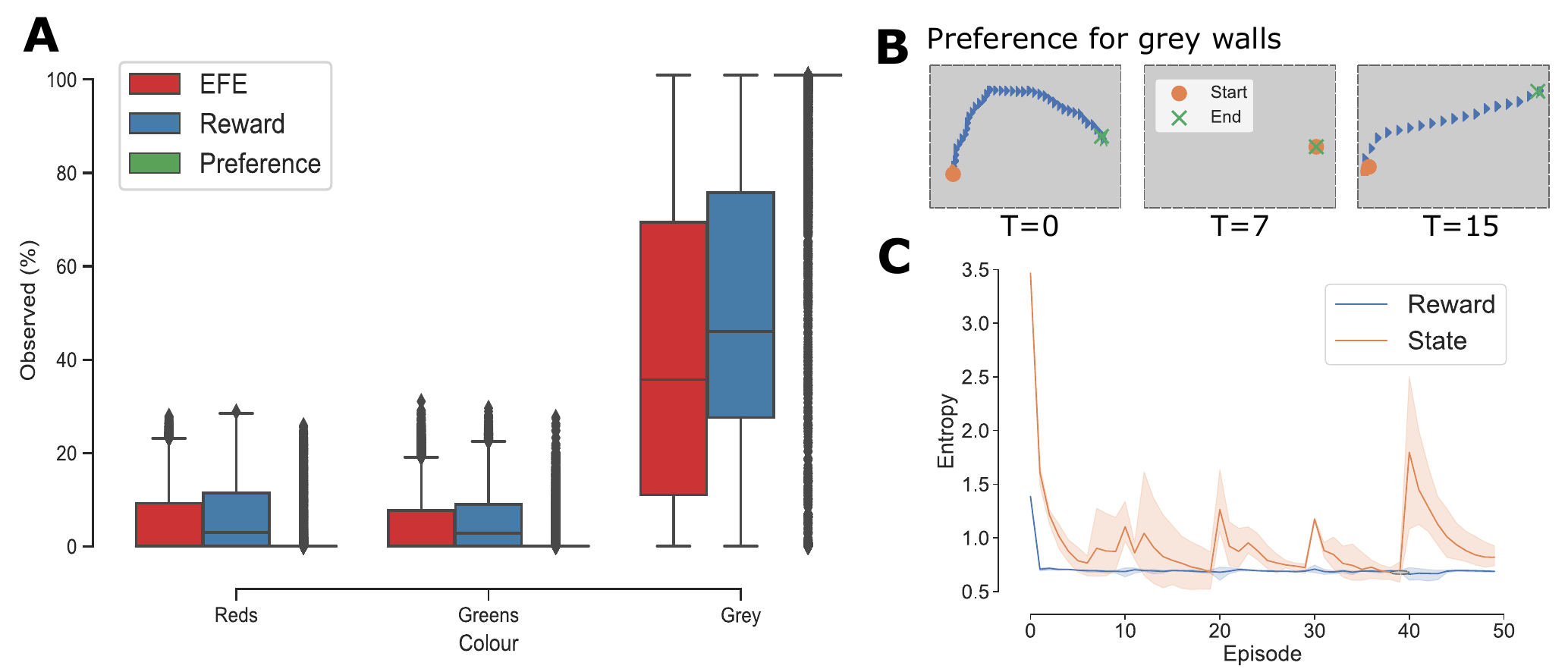}
  \caption{\textbf{A:} The bar chart plots the percentage of time the agents spent observing the $3$ colours in the TileWorld. The x-axis presents the colours: red (floor tile), green (floor tile) and grey (wall colour), and y-axis the percentage of observations calculated using the 32x32x3 pixel image the agent received at each time step. Red is for the agent optimising the standard expected free energy (EFE) Eq.\ref{e:efe}, blue for reward preference learning Eq.\ref{eq:G4} and green for state preference learning Eq.\ref{eq:G5}. \textbf{B:} Sample trajectories for a single agent are presented for agents acquiring a preference for observing grey walls during state preference learning. Here, an orange circle denotes starting position, a blue triangle represents the agents location until the final position, and a green cross is the final position. \textbf{C:} The line plot depicts the entropy over $P(s)$ (in orange) and $P(r)$ (in blue) across varying levels of volatility in the environment. The x-axis represents the episodes, and the y-axis entropy (in natural units). Here, the dark lines represent the mean (across $10$ seeds), and shaded area the $95\%$ confidence interval.}
  \label{fig:miniword} 
\end{figure}

Preference learning agents evinced a strong preference for looking at grey walls in the volatile TileWorld environment (Fig.\ref{fig:miniword}A). This was consistently observed for both preference learning over latent states and rewards. Importantly, when spawned in a location right next to the wall, these agents were happy to satisfy their preferences and not move  (Fig.\ref{fig:miniword}B). This is driven by three factors: fast preference accumulation over grey walls, a small number of state configurations and a generative model that is able to appropriately predict future trajectories (see example reconstructions and imagined roll-outs in the Appendix \ref{appendix::reconstructions}). However, volatility in the environment does influence the encoding of prior preferences -- as evident from the observed state entropy fluctuations across the episodes (Fig.\ref{fig:miniword}). 

\section{Concluding remarks}
\paragraph{Summary:} Pepper -- the reward-free preference learning mechanism presented -- provides a simple way to influence agent behaviour. Although, unlike the RL formulation, we cast \textit{what is preferred} to the agent instead of the environment 'designer'. That is, the agent is responsible for interacting with the environment and over time developing preferences that it acts to satisfy without an extrinsic signal. Our experiments revealed that rich category spaces are necessary for learning preferences that can establish distinct behavioural strategies -- specifically, within a volatile setting. Thus, future experiments looking to leverage \textit{Pepper} should provide a suitable category space to learn over.

\paragraph{Conjugate priors and Hebbian plasticity:} We employed a simple learning strategy for accruing preferences, using conjugate priors. This type of learning usually calls on associative or Hebbian plasticity \cite{hebb1949organization}, where synaptic efficacy is reinforced by the simultaneous firing of pre-and post-synaptic neurons. For example, as more neutral tiles are observed, more evidence is accumulated in the synaptic connection to support the hypothesis that neutral tiles are preferentially observed. Implicitly, Pepper rests upon experience-dependent plasticity, i.e., strengthening of synaptic connections during inference. These updates can have different times scales, and experiential levels (see Appendix \ref{appendix::plasticity}). 

Additionally, our learning process is driven by synaptic plasticity that allows certain random variables ($s,r$) to expand in light of new experiences \cite{fu2011experience}. They are only updated after an exchange with the environment. This separation of 'experiencing' the world, and then 'updating' model parameters is reminiscent of sleep; in which synaptic homeostasis resets model parameterisation, by encoding new synapses or removing redundant ones \cite{hinton1995wake,tononi2006sleep,hobson2012waking}. Removal of redundant model parameters is evaluated in experiments presented in Appendix \ref{appendix::plasticity}, where removal of accumulated Dirichlet parameters reveals consistently exploratory agents. 

\paragraph{Limitations:}
Notably, having a reward-free formulation is both an advantage and limitation of our approach. By not specifying an extrinsic reward function we forego control over the agent's behaviour. In other words, by removing the ability to manipulate agent preferences regarding what is considered rewarding leads to the removal of the only clear communication channel that can be used by a designer to control agent behaviour and/or define task goals. Therefore, it would not be appropriate to use pepper in a setting where control of the agent's behaviour is required e.g., for autonomous vehicles. And, if it were, precise preferences would have to be established under supervision.

Lastly, Pepper is contingent upon having a suitable Bayesian generative model that has all the necessary components for evaluating the EFE. Without this, Pepper may not work or the accumulated preferences might be misaligned with the observations. That is an inability to optimise the marginal likelihood over the random variables of interest. Therefore, great care should be placed to ensure that an appropriate model has been learnt (see Appendix \ref{appendix::reconstructions} for implementation details).


\paragraph*{Acknowledgements} We thank Fatima Sajid for reviewing the manuscript. NS acknowledges funding from the Medical Research Council, UK (MR/S502522/1). PT is supported by the UK EPSRC CDT in Autonomous Intelligent Machines and Systems (grant reference EP/L015897/1). KJF is funded by the Wellcome Trust (Ref: 203147/Z/16/Z and 205103/Z/16/Z).

\bibliography{paper}

\newpage
\appendix
\section{Pepper implementation}\label{appendix::efe}
Pepper was implemented as an extension to Dreamer V2~\citep{hafner2020mastering} public implementation\footnote{\url{https://github.com/danijar/dreamerv2}}. Specifically, Dreamer's generative model training loop was used, alongside a model predictive control (MPC) planner. Therefore, the actor learning part of Dreamer was not incorporated, and the generative model was trained using Plan2Explore~\cite{sekar2020planning}. Like Plan2Explore, an ensemble of image encoders were learnt and the ``disagreement'' of the encoders was used as an intrinsic reward during training. This guides the agent to explore areas of the map that have high novelty and potentially high information gain, when acquiring a generative (i.e., world) model. Unfortunately, replacing the amortised policy with this planner made the environment interaction (i.e., the acting loop) relatively slow. We implemented the planner as described in Algorithm~\ref{planner}.

\begin{algorithm}[H]
\label{planner}  
 \caption{Planner}
\SetEndCharOfAlgoLine{}
 \SetKwComment{Comment}{// }{}
 \SetKwInOut{Input}{Input}
 \Input{\\\hspace{-3.6em}\small
    \begin{tabular}[t]{l @{\hspace{.25em}} l}%
        $s_t$ & current state \\
        $N$ & Number of random action sequences to evaluate\\
    \end{tabular}%
  }
 \textbf{Initialise} \;
 \For{i=1\dots N}{
    $\pi^i \sim \mathcal{U}$ \tcc*[r]{Random action}
    $\text{score}_i \leftarrow 0$\;
    \For{$\tau=t\dots H$}{
        $\text{score}_i \leftarrow \text{score}_i - G(\pi^i, \tau)$ \tcc*[r] {Updated according Eq.\ref{eq:G4} or Eq.\ref{eq:G5}}
    }
 }
 $k \leftarrow \arg \max \text{score}$\;
 \textbf{Return} $\pi^k$
\end{algorithm}

Upon training completion, we froze the generative model learnt weights and only allowed learning of prior preferences. These (state or reward) preferences were updated after each episode, as described in Algorithm~\ref{pseudocode}.

\subsection{Evidence lower bound}
The generative model was optimised using the ELBO formulation introduced in \cite{hafner2019dream}:
\begin{align}\label{e:elbo_appendix}
     \mathcal{L}(\theta)=\sum\limits_{t=1}^T \big[&
     \underbrace{\underset{Q_\phi(s_t\mid o_{\leq t}, a_{\leq t})}{-\mathbb{E}\left[ \log P_\theta(o_t\mid s_t,\pi) \right]} - \underset{Q_\phi(s_t\mid o_{\leq t}, a_{\leq t})}{\mathbb{E}\left[\log P_\theta(r_t\mid s_t,\pi) \right]}}_{\mbox{reconstruction}}\\
     + & \underbrace{\underset{Q_\phi( s_t\mid o_{\leq t}, a_{\leq t} )\qquad \qquad\qquad\qquad\qquad}{\mathbb{E}\left[ D_{KL}(Q_\phi(s_t\mid o_t, s_{t-1}, \pi)) \parallel P_\theta(s_t\mid s_{t-1},\pi) \right]}}_{\mbox{dynamics}} \big]\nonumber ~ .
     \vspace{-1em}
\end{align}

\subsection{Expected free energy for pepper}
We implemented EFE using the parameterisations introduced in \citep{fountas2020deep} and adapted for \citep{hafner2020mastering}:

\begin{itemize}
    \item \textbf{Term~\ref{eq:G4a}} was modelled as a categorical likelihood model (using normalised Dirichlet counts).
    \item \textbf{Term~\ref{eq:G4b}} was computed as the $KL$ divergence between the prior ($s \sim Q(s_\tau | \pi)$) and the posterior ($s \sim P(s_\tau|o_\tau,\pi)$) states. This could be computed analytically because the prior and posterior state distributions were modelled as Categorical distributions. Here, the dependency on the policy $\pi$ was accounted for by using the RNN hidden state $h_t$ summarising past actions and roll-outs.
    
    \item \textbf{Term~\ref{eq:G5a}} was computed as the entropy of the observation model $P(o_\tau|s_\tau,\pi)$. Happily, the factorisation of the observation model -- as independent Gaussian distributions -- allowed us to calculate the entropy term in closed form.
    
    \item \textbf{Term~\ref{eq:G5b}} was computed as the difference between $\log Q(s_\tau|\pi)$ and $\log P(s_\tau|D)$, where $\log Q(s_\tau|\pi)$ was approximated using a single sample from the prior model $Q(s_\tau|\theta,\pi)$. Again, the dependency on $\pi$ was substituted by $h_t$.
    
    \item \textbf{Terms~\ref{eq:G4c} and~\ref{eq:G5c}} were more challenging to compute. Like \citep{fountas2020deep}, we rearranged the expression to $H(o_\tau|s_\tau,\theta,\pi)-H(o_\tau|s_\tau,\pi)$. This translates to $I(o_\tau;\theta|s_\tau,\pi)$, and can be approximated using Deep Ensembles~\cite{lakshminarayanan2016simple,sekar2020planning} and calculating their variance $\text{Var}_\theta[\mathbb{E}Q(o_\tau|s_\tau,\theta,\pi)]$. Here, each ensemble component can be seen as a sample from the posterior $Q(\theta|s_\tau,\pi)$. Our experiments showed that using 5 components was sufficient.

\end{itemize}

\section{Experiments}\label{appendix::traingenmodel}

\paragraph{FrozenLake} 
For these experiments, we simulated the agent in five distinct situations ranging from a non-volatile, static environment to a highly volatile one i.e., a different FrozenLake map every step. For all episodes in the static setting, the agent was initialised at a fixed location with no changes to the FrozenLake map throughout that particular episode. Conversely, agents operating in the volatile setting were initialised at a different location each time. Moreover, the FrozenLake map was also changed every $N$ steps -- given the desired volatility level. For $100\%$ volatility the map changed every step, $75\%$ volatility corresponded to map changes every $10$ steps, $50\%$ volatility corresponded to map changes every $20$ steps and $25\%$ volatility corresponded to map changes every $40$ steps. Additionally, the generative model used for these experiments was trained in a volatile setting where the map changed every $5$ steps (Table \ref{table:training_param}).

\paragraph{TileWorld}
For these experiments, we simulated the agent under two conditions (non-volatile and volatile). For the non-volatile setting, the agent was initialised at a fixed location with no changes to the TileWorld map throughout training and testing. In the volatile setting, for every $K$ step, we toggle  alternate between a randomly sampled map and the original map. This allowed us to simulate uncertain states that trigger exploratory behaviour. The generative model used for these experiments was trained in a volatile setting where the map changed every $10$ steps (Table \ref{table:training_param}).

\begin{table}[h!]
\caption{Training parameters}
\centering
\begin{tabular}{l|l|l}
 \textbf{Parameter}&\textbf{FrozenLake}&\textbf{TileWorld}\\
\midrule
 Planning Horizon & $15$ steps & $15$ steps \\
 Episode Length & $50$ steps  & $200$ steps \\
 Reset Every & $5$ steps & $10$ steps \\
 No. Episodes & $50$ episodes & $50$ episodes \\
 No. State Categories & $64$ categories & $32$ categories \\
 No. State Dimensions & $50$ dimensions & $50$ dimensions \\
 No. Reward Categories & $4$ categories & $3$ categories \\
\end{tabular}\label{table:training_param}
\end{table}

\begin{table}[h!]
\caption{Preference learning parameters for long-term learning}
\centering
\begin{tabular}{l|l|l}
 \textbf{Parameter}&\textbf{FrozenLake}&\textbf{TileWorld}\\
\midrule
 Planning Horizon & $15$ steps & $15$ steps\\
 Episode Length & $50$ steps & $200$ steps\\
 No. Episodes & $50$ episodes & $50$ episodes\\
 Reset Map Every & $1,10,20,40,50$ steps  & $10$ steps\\
 No. Agents & $10$ agents & $3$ agents\\
\end{tabular}
\end{table}

\subsection{Computational requirements}
Overall, our experiments required $1344$ GPU hours. Each GPU was a GeForce RTX $3090$.

\subsection{Image reconstruction and imagined roll-outs}\label{appendix::reconstructions}
\begin{figure}[ht]
  \centering
  \includegraphics[width=\linewidth]{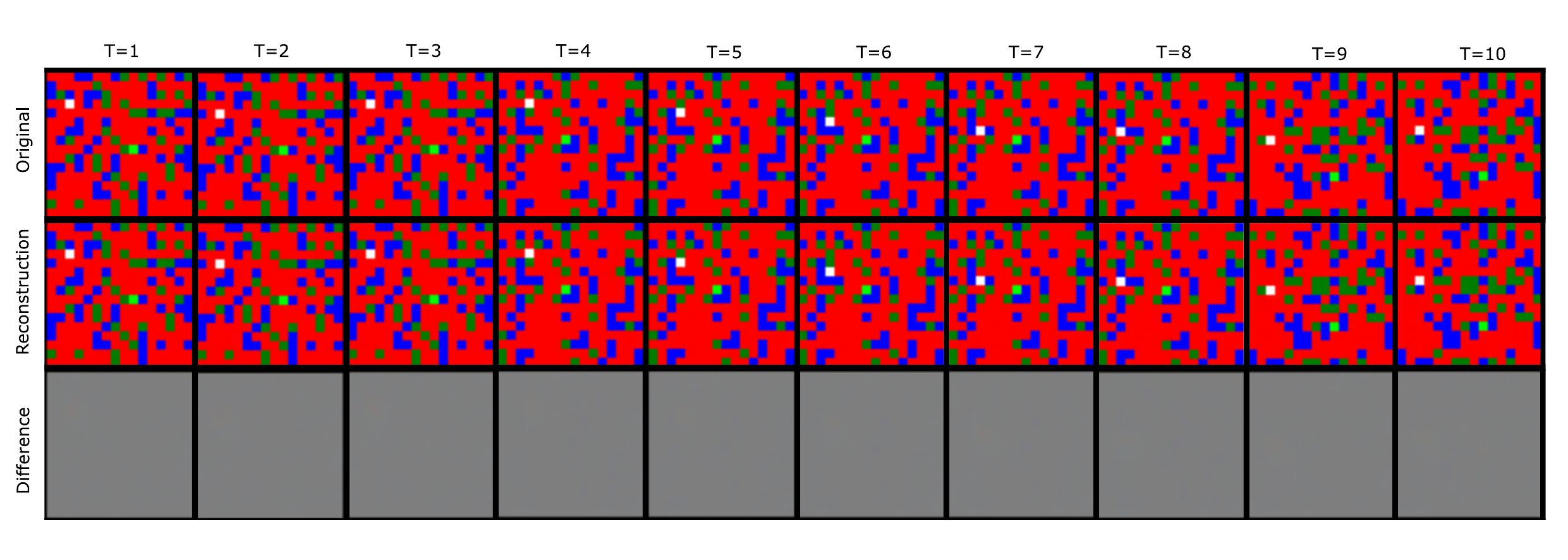}
  \caption{An example of the FrozenLake reconstruction for the first $10$ steps of an episode, with map changes at $T=4~ \&~ 9$.}
  \label{fig:reconstruction_fl} 
\end{figure}

\begin{figure}[ht]
  \centering
  \includegraphics[width=\linewidth]{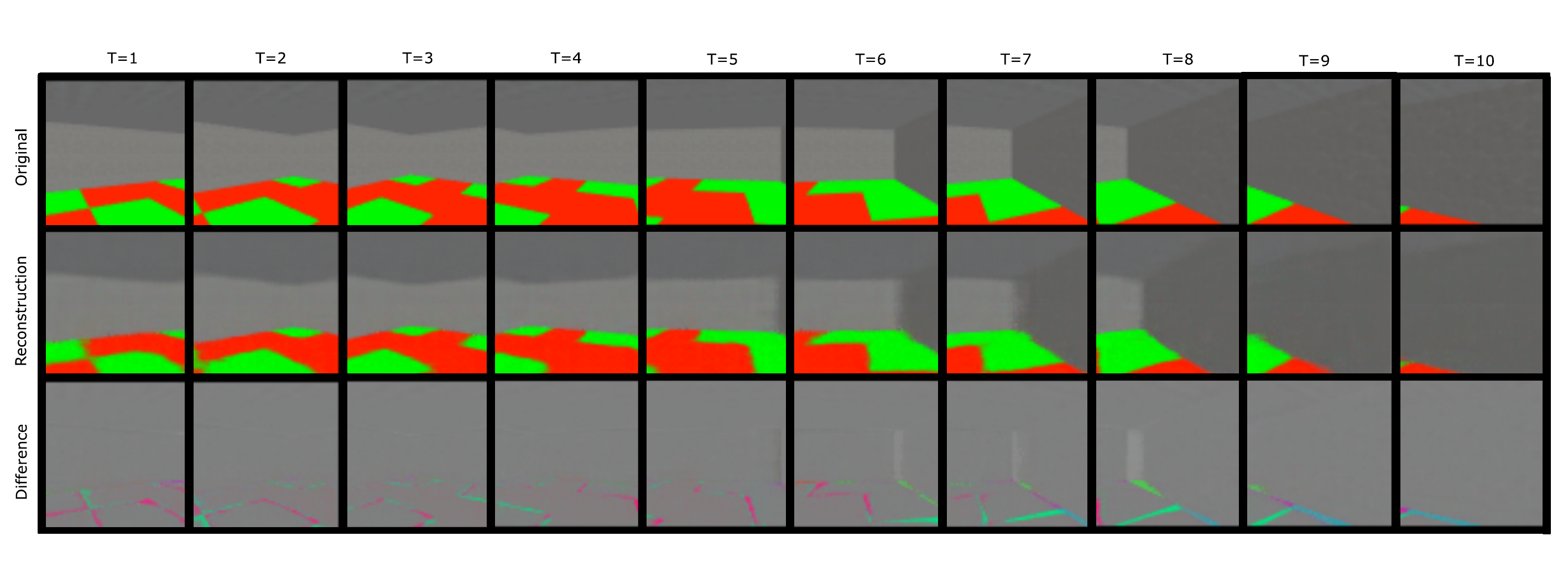}
  \caption{An example of TileWorld reconstruction for the first $10$ steps of an episode.}
  \label{fig:reconstruction_tw} 
\end{figure}

\begin{figure}[ht]
  \centering
  \includegraphics[width=\linewidth]{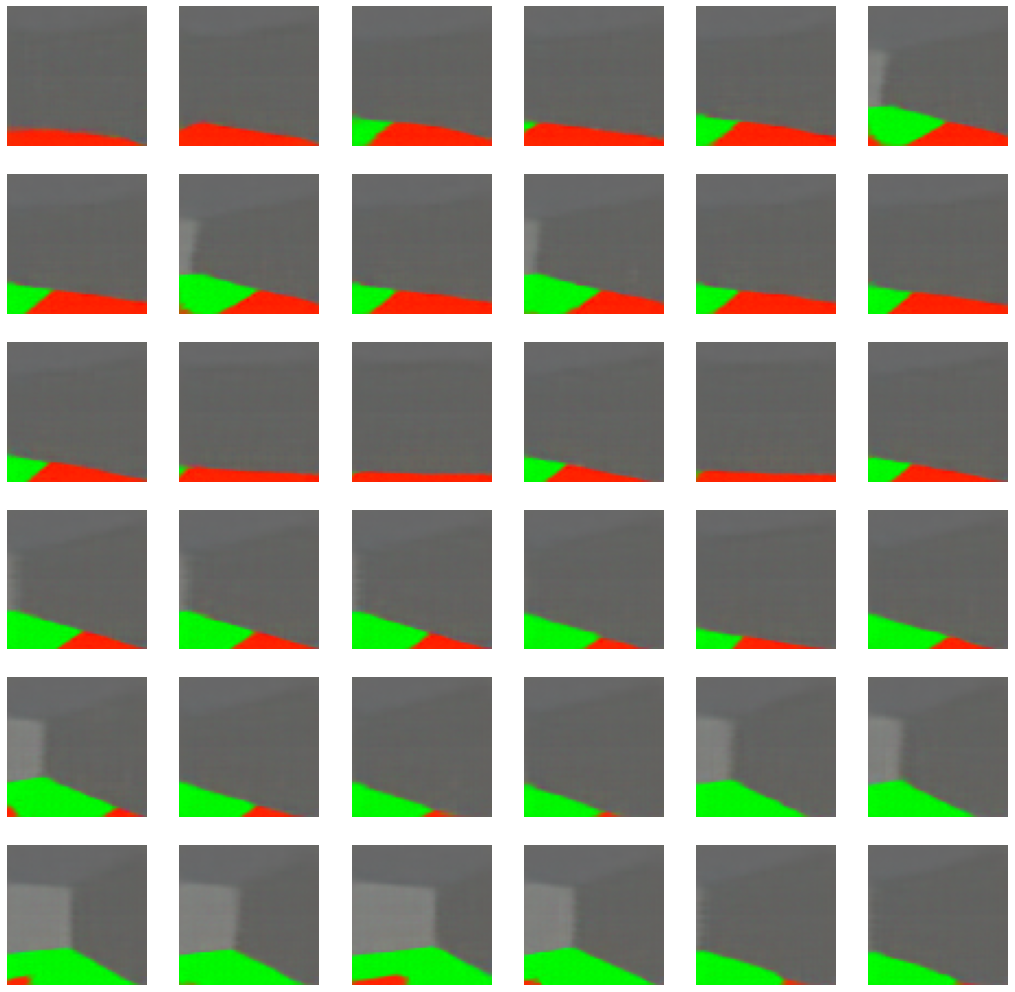}
  \caption{An example the observations from an imagined roll-out in the latent state space. These roll-outs are constructed using a random action trajectory that is propagated forward to get a  latent state sequence. For each of the latent states the observations were sampled from the observation model $P(o_t|s_t)$.}
  \label{fig:tileworld_rollout} 
\end{figure}

For apt learning of preferences, the agent's generative model must be able to accurately infer the current (and future) states of affairs. To evaluate this for our learnt generative models, we illustrate representative examples of reconstructions encoded by the pepper agents for a particular episode. Fig.\ref{fig:reconstruction_fl} shows the reconstructions for FrozenLake, and Fig.\ref{fig:reconstruction_tw} for TileWorld. The imagined roll-outs for the TileWorld environment are shown in Fig.\ref{fig:tileworld_rollout}. 

\section{Behaviour under long-term preference learning}

\subsection{Learnt preferences}\label{appendix::preferences}
\begin{figure}[!ht]
  \centering
  \includegraphics[width=\linewidth]{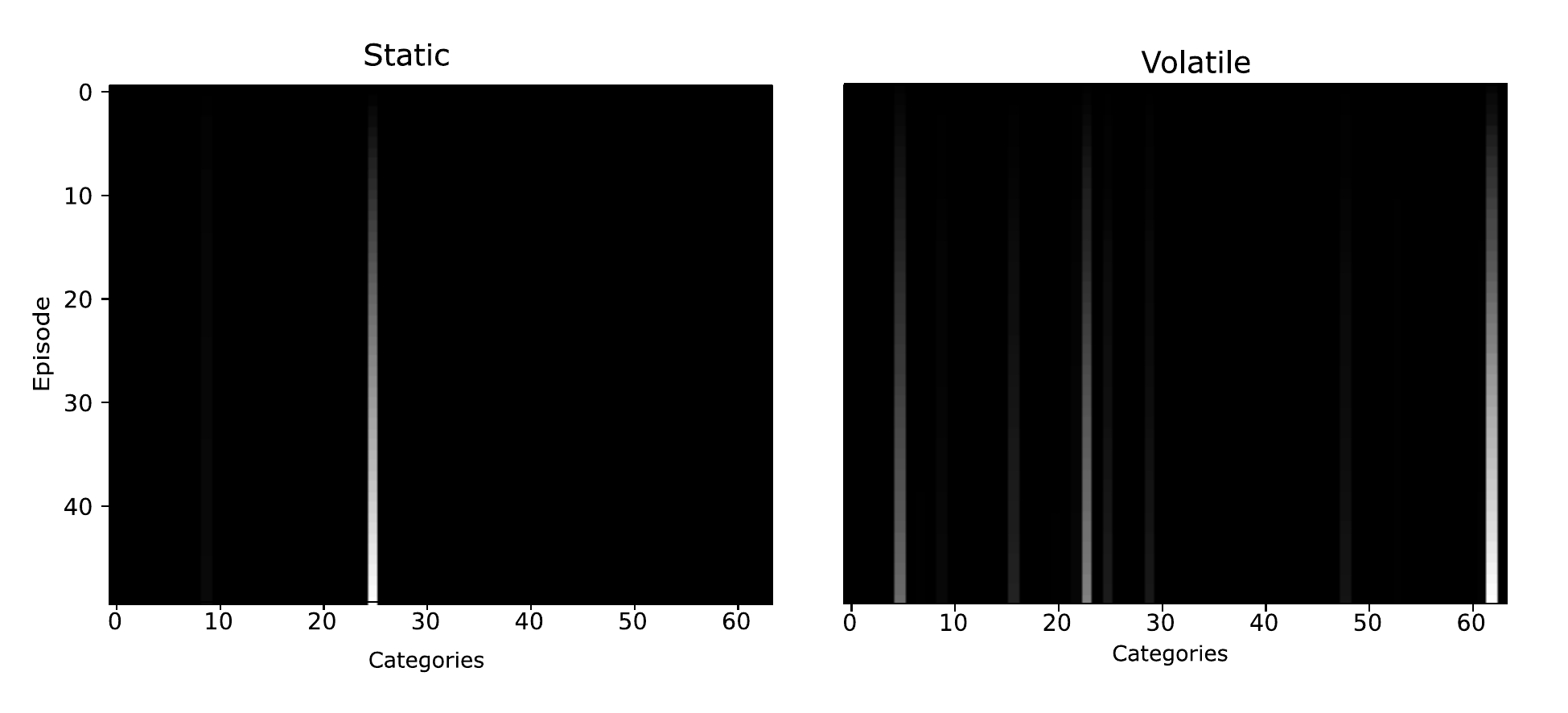}
  \caption{An example of learnt state preferences for a single agent in a static and highly volatile setting. Here, $64$ state categories are presented on the x-axis and episodes on the y-axis. The first panel is for preferences learnt under a static setting, and the second for preferences learnt under a volatile setting. The scale goes from white (i.e., high Dirichlet concentration) to black (i.e., low Dirichlet concentration), and grey indicates gradations between these.}
  \label{fig:pref_state} 
\end{figure}

\begin{figure}[!ht]
  \centering
  \includegraphics[width=\linewidth]{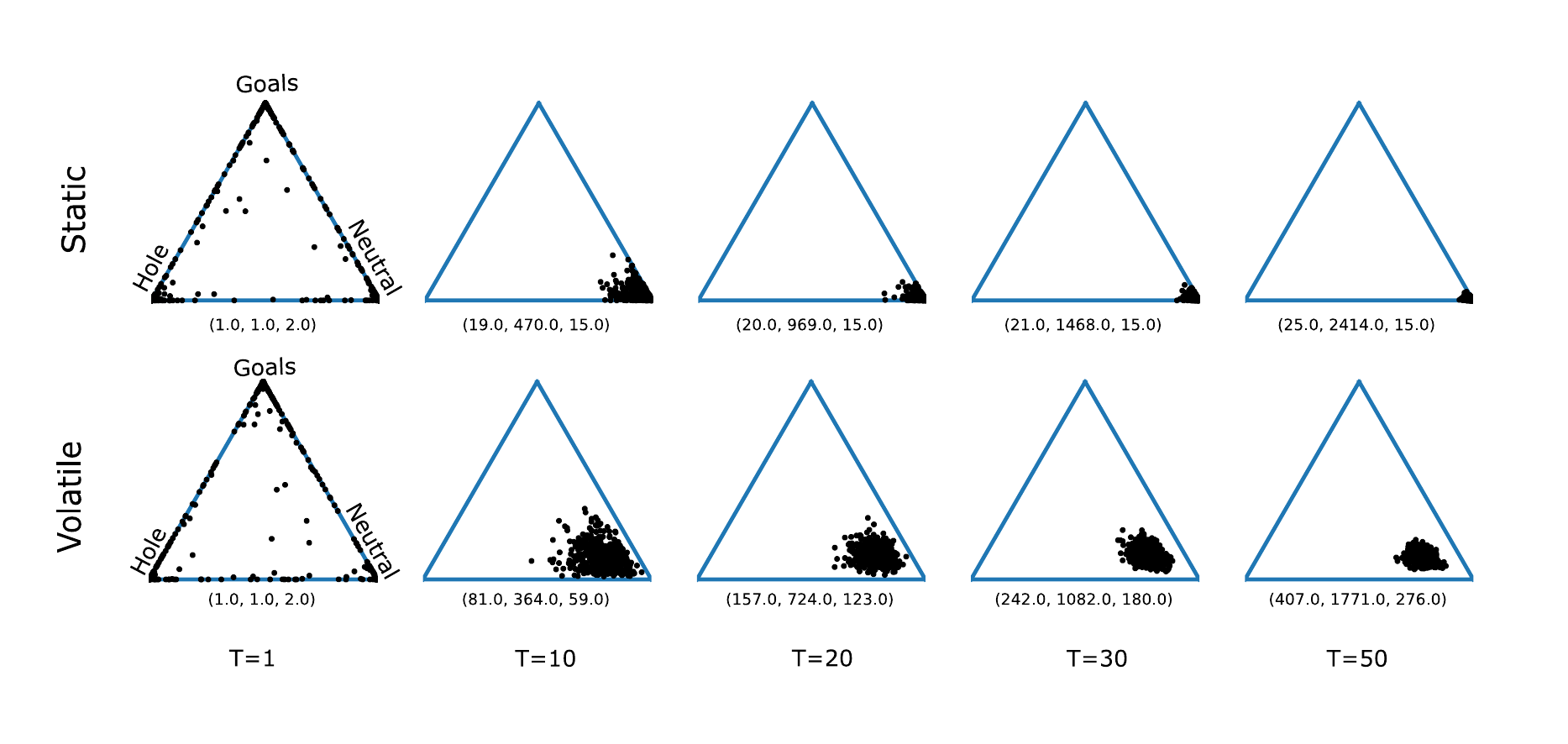}
  \caption{An example of learnt reward preferences for a single agent in a static and highly volatile setting. The first row is for preferences learnt under a static setting, and the second row for preferences learnt in a volatile setting. Each figure illustrates the Dirichlet distribution in a 3-dimensional coordinate space, i.e., 2-simplex -- for a particular episode ($T$). Here, the concentration of dots in one corner reflect precise beliefs; and scattered dots denote imprecise beliefs. Each dot represents a single sample from the Dirichlet distribution (determined by the alpha parameters denoted at the bottom of each figure), and each plot displays $500$ samples. For clarity, we collapsed Goal and Sub-goals into one category. Preferences for both static and volatile setting are initialised as uniform (i.e., $(1,1,1,1) = (1,1,2)$) denoted by the dots scattered across the simplex.}
  \label{fig:pref_reward} 
\end{figure}

We expected differences in learnt preferences to induce shifts in agent behaviour. For example, agents who repeatedly accrued Dirichlet pseudo-counts for the same category would exhibit preference satisfying behaviour. This would be due to high precision (or confidence) over that particular category. In contrast, an agent who accrued Dirichlet pseudo-counts for different categories would exhibit exploratory behaviour given an imprecise (or low confidence) distribution over the categories. To illustrate how different environment settings shaped preference learning we looked at the static and volatile setting where the FrozenLake map changed every step. Fig.\ref{fig:pref_state} shows a representative example of state preference learning under these conditions, and Fig.\ref{fig:pref_reward} an example of reward preference learning. We observed that state preferences learnt under a static setting were precise -- denoted by the repeated pseudo-count accumulation over category $25$. Conversely, for the volatile setting an imprecise state preference distribution was learnt (Fig.\ref{fig:pref_state}). Separately, we observed that the learnt reward preferences were precise -- regardless of the setting. We posit that this is a consequence of differences in the reward and state category space. In other words, having a large number of state categories allowed distinct preferences to be learnt under static and volatile settings. This is reflected in the qualitative differences seen between the two (Fig.\ref{fig:pref_state}). 

\subsection{Agent trajectories}\label{appendix::exploration}
\begin{figure}[ht]
  \centering
  \includegraphics[width=\linewidth]{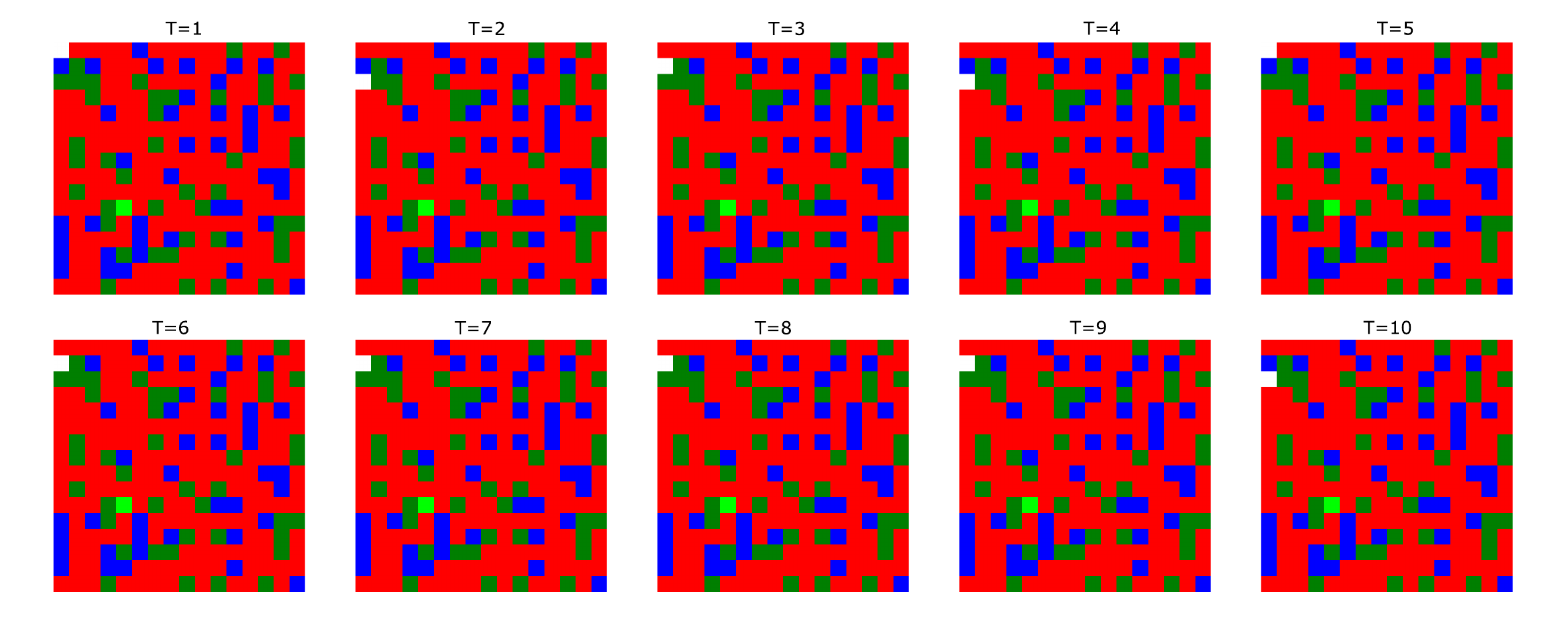}
  \caption{Representative example of the agent trajectories observed during state preference learning under a static setting.}
  \label{fig:static_traj} 
\end{figure}

\begin{figure}[ht]
  \centering
  \includegraphics[width=\linewidth]{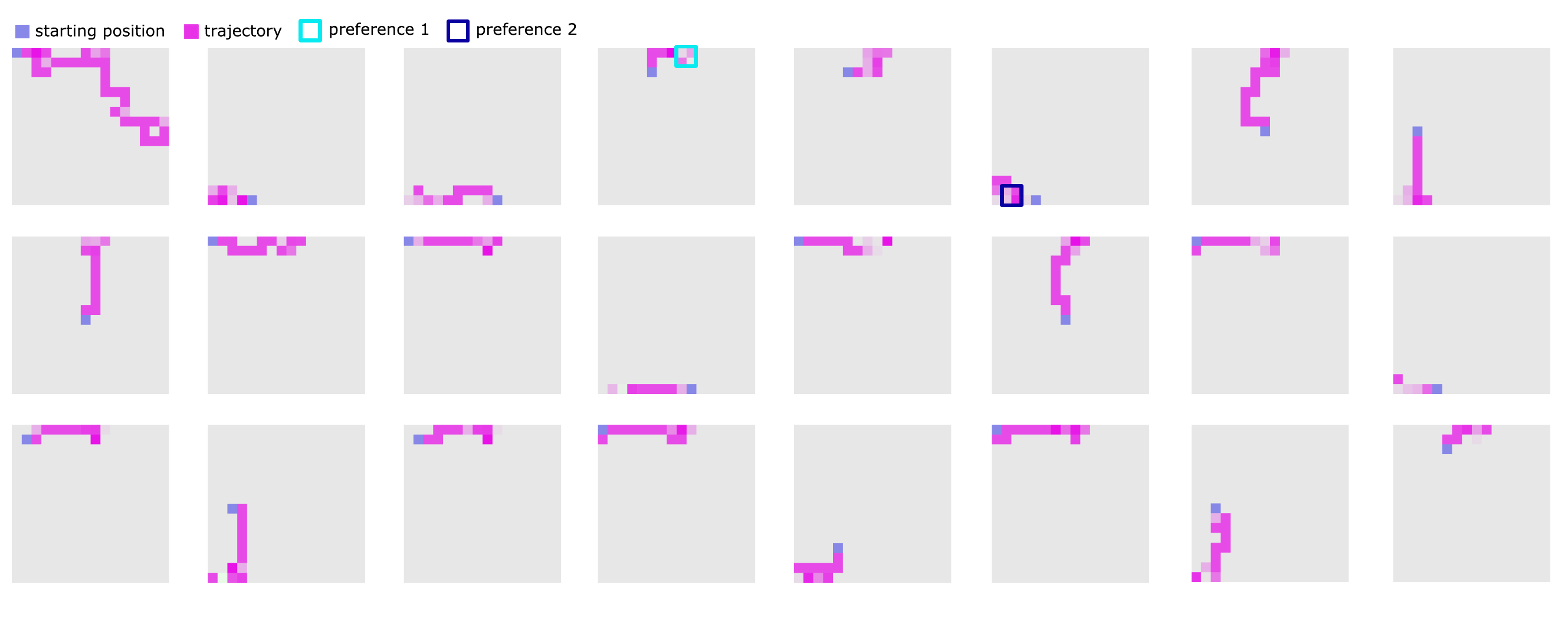}
  \caption{Representative example of the agent trajectories observed during state preference learning under the volatile setting. Each figure is an illustration of the agents trajectory for a particular episode. Here, purple is the agents starting position, pink the trajectory, cyan square denotes the first learnt preference and dark blue denotes the second learnt preference.}
  \label{fig:volatile_traj} 
\end{figure}

Next, we evaluated how disparate the agent trajectories were given the observed differences in preference accumulation (Figure \ref{fig:pref_state} and \ref{fig:pref_reward}). For the static setting, we observed agents satisfying their preferences by restricting movement to a small patch in the FrozenLake. This behaviour was observed consistently across all agents (i.e., different seeds) and episodes. We present a representative example in Fig.\ref{fig:static_traj}. Separately, agents simulated in the volatile setting (where the map changed every step) learnt a bi-modal preference set (i.e., preferred to go to one of two locations in the FrozenLake). Here, the location preference depended on the initial location i.e., if the agent was initialised in a tile close to the first preferred location then it choose to go there. However, the second location was preferred if the agent was initialised close to it. We present a representative example in Fig.\ref{fig:volatile_traj}. Interestingly, this behaviour was observed in agents where the environment was $100\%$ volatile, whereas agents operating in slightly less volatile settings continued exploring (Fig.\ref{fig:tradeoff}A). 

\begin{figure}[ht]
  \centering
  \includegraphics[width=\linewidth]{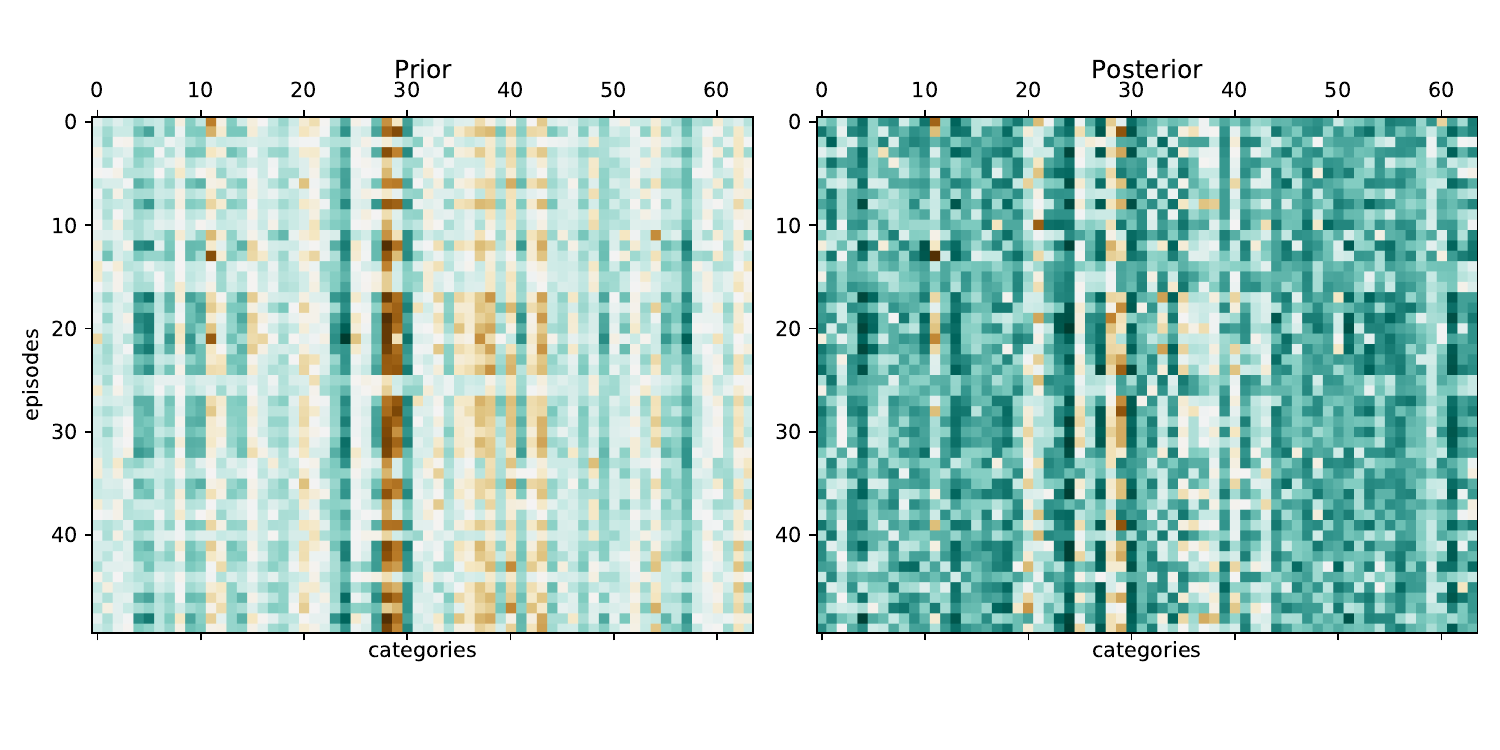}
  \caption{Variance over the estimated posterior  ($s \sim Q_\phi(s_t|h_t,o_t)$) and priors ($s \sim Q_\phi(s_t|h_t)$) under the volatile setting.  The scale goes from brown (low variance) to turquoise (high variance), and light shades indicate gradations between these.
  }
  \label{fig:variance} 
\end{figure}

Importantly, these agents were able to disregard, noisy information about the states from the environment. To qualify this, we looked at how the variance between the posterior ($s \sim Q_\phi(s_t|h_t,o_t)$) and prior ($s \sim Q_\phi(s_t|h_t)$) estimates differed across the $50$ episodes for these agents (Fig.\ref{fig:variance}). We observed that the posterior estimates had a greater variance across the $50$ dimensions relative to the prior variance. Given how these estimates are calculated, we postulate that differences in the variance were due to the change in the FrozenLake map that the agent finds itself in after it moved one step. These high variances in the posterior estimate, under a highly volatile setting, induced a change in behaviour from exploratory to preference satisfaction. 

\section{Behaviour under short-term preference learning}\label{appendix::plasticity}
We expected reduced preference learning timescales to influence the agent's preferred behaviour. To evaluate this, we consider a setting where the agent was equipped with a sliding preference window i.e., after $k$ steps the previous preferences were removed in favour of new ones. To evaluate short-term preference learning, we considered state preferences with a sliding window of $5$ episodes (Table \ref{table:short_term}). 

\begin{table}[h!]
\caption{Preference learning parameters for short-term learning}
\centering
\begin{tabular}{l|l}
 \textbf{Parameter}&\textbf{FrozenLake}\\
\midrule
 Planning Horizon & $15$ steps \\
 Episode Length & $50$ steps\\
 No. Episodes & $50$ episodes \\
 Reset Map Every & $1,10,20,40,50$ steps \\
 Reset Preference Every & $5$ episodes \\
 No. Agents & $5$ agents \\
\end{tabular}\label{table:short_term}
\end{table}

\begin{figure}[ht]
  \centering
  \includegraphics[width=\linewidth]{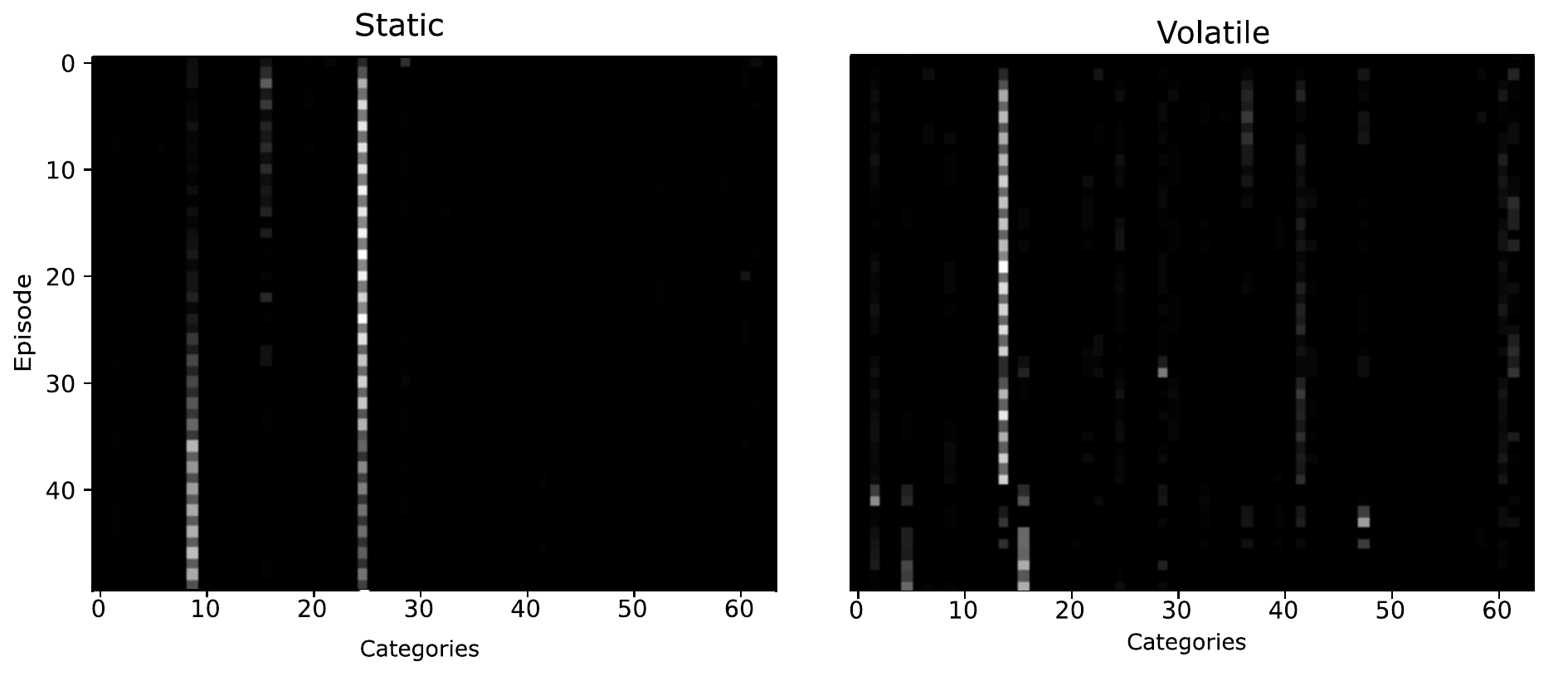}
  \caption{An example of short term state preference learning for an agent in a static and highly volatile (map change every step) environment. Here, $64$ state categories are presented on the x-axis and episodes on the y-axis. The first panel is for preferences learnt under a static setting, and the second for preferences learnt in a volatile setting. The scale goes from white (high Dirichlet concentration) to black (low Dirichlet concentration), and grey indicates gradations between these.}
  \label{fig:pref_st_state} 
\end{figure}

\begin{figure}[!htp]
  \centering
  \includegraphics[width=\linewidth]{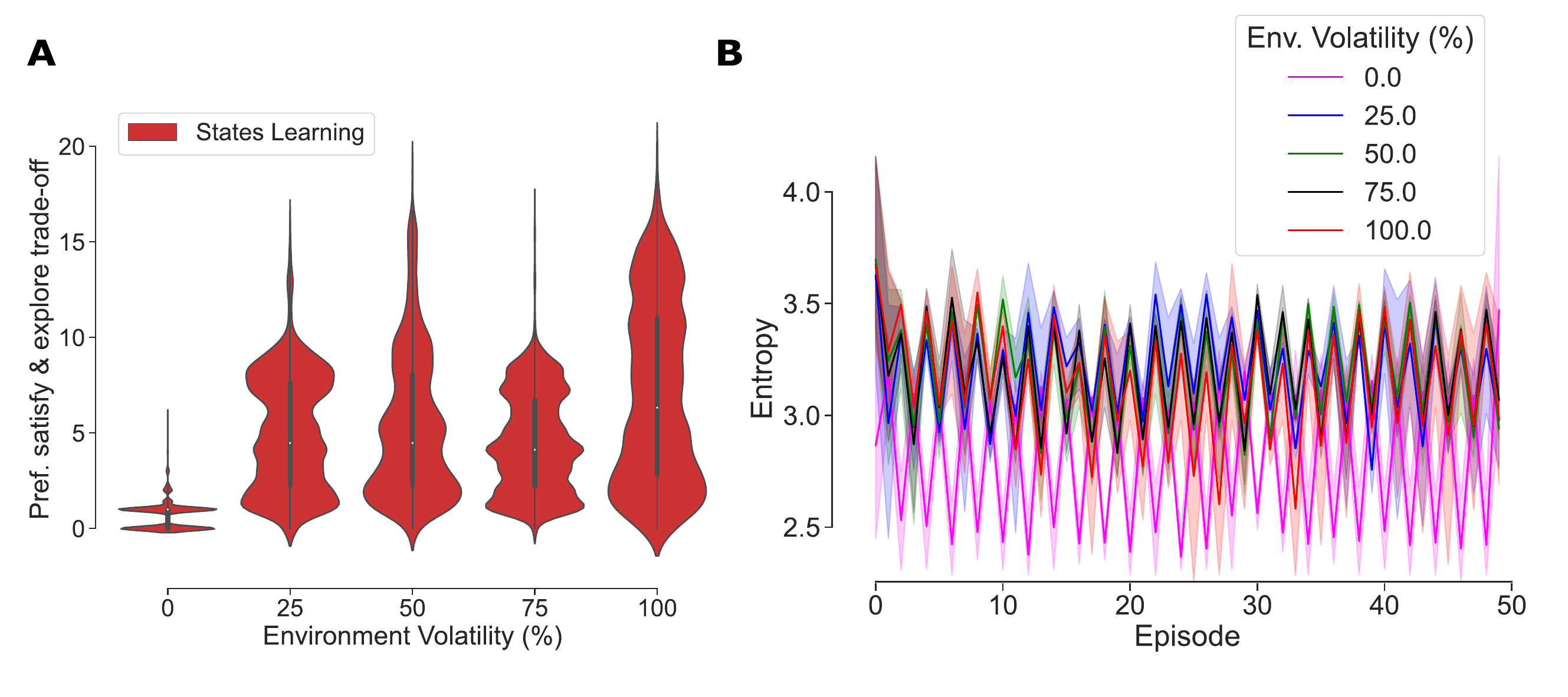}
  \caption{\textbf{A:} The violin plot illustrates the preference satisfaction and exploration trade-off measured using the Hausdorff distance \cite{blumberg1920hausdorff} at different levels of volatility in the environment, when the agent had short term preferences. The x-axis denotes environment volatility: with a constant map ($0\%$), change in map every $40$ steps ($25\%$), $20$ steps ($50\%$), $10$ steps ($75\%$) and every step ($100\%$). The y-axis denotes the Hausdorff distance. Here, red is for the agent optimising state preference learning Eq.\ref{eq:G5}. \textbf{B:} The line plot depicts the entropy over $P(s)$ across varying levels of volatility in the environment. The x-axis represents the episodes, and the y-axis entropy (in natural units). Here, the dark lines represent the mean (across $5$ seeds), and shaded area the $95\%$ confidence interval. The pink line is for $0\%$, blue for $25\%$, green for $50\%$, black for $75\%$ and red for $100\%$ volatility in the environment. The spikes in entropy correspond to overwriting of learnt preferences.}
  \label{fig:tradeoff-st} 
\end{figure}

For this, we looked at the preferences learnt in the static and volatile (map changes every $10$ steps) setting (Fig.\ref{fig:pref_st_state}). Predictably, we observed differences in the preference accumulation when the agents learnt short-term preferences regardless of the setting. Explicitly, in the static setting the accrued preferences were flexibly learnt and unlearnt over time e.g., category $24$ was slowly unlearnt in favour of $11$ category. Volatile conditions fostered perpetual preference uncertainty as accumulated Dirichlet pseudo-counts were repeatedly updated. Therefore, we would expect these agents to exhibit exploratory behaviour compared to agents equipped with long-term preferences due to imprecise preference learning. To quantify this behaviour, we projected the latent states onto the first two components (fitted using long-term state preferences simulation data). The short-term simulation data only mapped onto a small space in Fig.\ref{fig:learnpreferences} latent space. Furthermore, there was no clear separation between projected latent states across the the volatile and static settings. This is reflected in the increased state entropy ($\sim 0.5$ nats),  under both settings, as previously learnt preferences were removed (Fig.\ref{fig:tradeoff-st}). 

Next, we considered how the exploration and preference satisfaction trade-off might vary when preferences were learnt over a short time horizon. Using the Hausdorff distance, we evaluated how the volatility in the environment changed the agent's behaviour to either exploratory or satisfying preferences (Fig.\ref{fig:tradeoff-st}). In contrast to the long-term state preference learning setting, we see a (slight) linear association between environment volatility and preference satisfaction. This is consistent with our expectation that a slow removal of accumulated Dirichlet parameters engenders consistently exploratory agents.

\begin{figure}[!tp]
  \centering
  \includegraphics[width=0.8\linewidth]{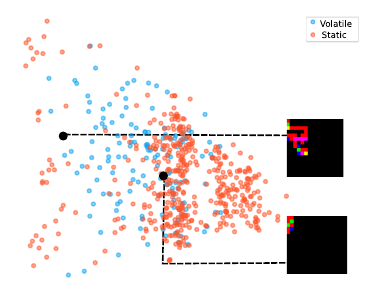}
  \caption{Visualisation of the posterior latent states (estimated using $Q_\phi(s_t|h_t,o_t)$) during state preference learning. The states have been projected onto the first two principle components (fitted using long-term state preferences simulation data), and the black circles represent their k-mean centroid. Here, the accompanying graphics present a representative agent trajectory with visited tiles highlighted from that particular cluster. }
  \label{fig:stminiword} 
\end{figure}

\end{document}